\documentclass[conference]{IEEEtran}
\IEEEoverridecommandlockouts
\usepackage[para]{footmisc}
\usepackage{amsmath,amssymb,amsfonts}
\usepackage{algorithmic}
\usepackage{graphicx}
\usepackage{textcomp}
\usepackage{svg,url}
\usepackage[colorlinks=true, allcolors=blue]{hyperref}
\usepackage{cite,cleveref}
\def\BibTeX{{\rm B\kern-.05em{\sc i\kern-.025em b}\kern-.08em
    T\kern-.1667em\lower.7ex\hbox{E}\kern-.125emX}}

\usepackage[pscoord]{eso-pic}%
\newcommand{\placetextbox}[3]{
  \setbox0=\hbox{#3}
  \AddToShipoutPictureFG*{
    \put(\LenToUnit{#1\paperwidth},\LenToUnit{#2\paperheight}){\vtop{{\null}\makebox[0pt][c]{#3}}}%
  }%
}%
\begin{document}
\title{Deep Learning Mental Health Dialogue System}

\author{\IEEEauthorblockN{Lennart Brocki}
\IEEEauthorblockA{\textit{Institute of Informatics} \\
\textit{University of Warsaw}\\
Warsaw, Poland \\
brocki.lennart@gmail.com}
\and
\IEEEauthorblockN{George C. Dyer}
\IEEEauthorblockA{
\textit{Demiteris}\\
Wrocław, Poland \\
georgecdyer@gmail.com}
\and
\IEEEauthorblockN{Anna Gładka}
\IEEEauthorblockA{\textit{Psychiatry Department} \\
\textit{Wrocław Medical University}\\
Wrocław, Poland\\
agladka@gmail.com}
\and
\IEEEauthorblockN{Neo Christopher Chung}
\IEEEauthorblockA{\textit{Institute of Informatics} \\
\textit{University of Warsaw}\\
Warsaw, Poland \\
nchchung@gmail.com}
}

\maketitle 
\placetextbox{.5}{.99}{\large{6th International Workshop on Dialog Systems (IWDS)}}
\placetextbox{.5}{0.97}{\large{10th IEEE International Conference on Big Data and Smart Computing (BigComp)}}

\begin{abstract}
Mental health counseling remains a major challenge in modern society due to cost, stigma, fear, and unavailability. We posit that generative artificial intelligence (AI) models designed for mental health counseling could help improve outcomes by lowering barriers to access. To this end, we have developed a deep learning (DL) dialogue system called \emph{Serena}. The system consists of a core generative model and post-processing algorithms. The core generative model is a 2.7 billion parameter Seq2Seq Transformer \cite{vaswani2017attention} fine-tuned on thousands of transcripts of person-centered-therapy (PCT) sessions. The series of post-processing algorithms detects contradictions, improves coherency, and removes repetitive answers. Serena is implemented and deployed on \url{https://serena.chat}, which currently offers limited free services. While the dialogue system is capable of responding in a qualitatively empathetic and engaging manner, occasionally it displays hallucination and long-term incoherence. Overall, we demonstrate that a deep learning mental health dialogue system has the potential to provide a low-cost and effective complement to traditional human counselors with less barriers to access. \textbf{}
\end{abstract}


\section{Introduction}

The lack of widespread access to mental health counseling remains one of the biggest challenges in the world. It is estimated that 658 million people in the world suffer from some form of psychological distress and this number grew by 50\% in the last 30 years \cite{gbd2022global}. Yet only 35\% percent of people with mental health disorders receive mental health treatment \cite{chen2021accessibility}, and less than 25\% percent have ever ``seen someone'' \cite{engels2020depression}. Psychological counseling and therapy are helpful in treating anxiety, depression, obsessive compulsive disorder, personality disorders, eating disorders and a plethora of other conditions \cite{lazar2014cost}. Around 48\% of people experiencing a mental health crisis reported that talking with friends was helpful, however 56\% of them ended up handling their problems alone \cite{ebert2019barriers}. We propose that a virtual mental health counselor based on generative deep learning models could substantially improve mental health outcomes for many user profiles. In this paper we will present our design and implementation of a deep learning dialogue system for psychological counseling.

Generative deep learning (DL) models may provide an answer to a simple yet tenacious question: how can we make mental health counseling more accessible? To effectively tackle the problem, we first need to consider why most people cannot or \textit{do not want to} access mental health counseling. The most obvious cause is the prohibitive cost of the type of regular, in-person counseling that is proven to be the most beneficial \cite{hofmann2012efficacy}. A similar obstacle is time. Those people who earn enough money to afford quality counseling may not, as a result, have enough time to dedicate to the process, which in addition to the actual sessions, requires scheduling, commuting, arranging for the care of children, etc. Finally, we have fear of counseling and perceived stigma \cite{prior2012overcoming}. 

We designed such a DL-based dialogue system called \emph{Serena} as a system that addresses as many of these factors as possible, with an emphasis on filling the gaps left by traditional, in-person counseling. Stated differently, the proposed system is not designed as a replacement for traditional therapy. Rather, we conceive it as: 1) a fallback for those who are strictly unable to engage in traditional therapy because of money or time; 2) a catalyst for helping people warm up to the idea of sharing their thoughts through the process of dialogue, which may result in them in setting up in-person sessions; 3) a tool for identifying therapy needs and measuring engagement with a virtual counseling model across a wide demographic, with the goal of improving quality and access to mental health resources globally.

\section{Related Work}

\begin{figure}[bt!]
\centering 
\includegraphics[width=0.5\textwidth]{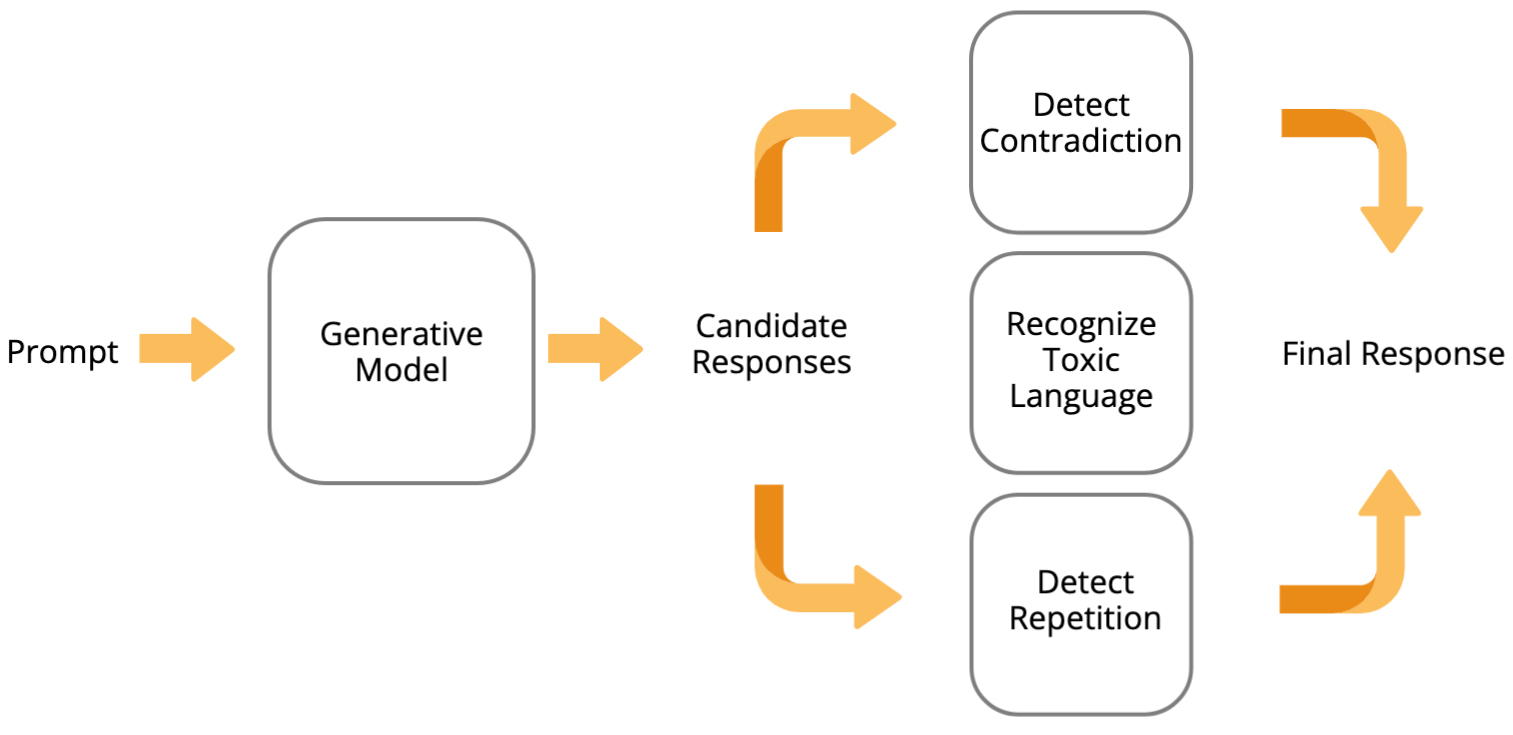}
\caption{Overview of the Serena dialogue system. A large generative model outputs candidate responses conditioned on a user prompt and three smaller, more specialized NLP models are used to reject  unsuitable responses.} 
\label{overview}
\end{figure}

Broadly speaking, dialogue systems (a.k.a. chatbots) can be divided into two groups: those that primarily use artificial intelligence or generative processes on the one hand, and those that primarily use symbolic methods or hard coding on the other. The use of virtual dialogue systems in the context of mental health counseling dates backs to the unveiling of the ELIZA program in 1964 \cite{weizenbaum1966eliza}, which used symbolic methods to deterministically process inputs and generate responses. Contemporary efforts such as Wysa, Woebot, and Joy use machine learning to process the user's input and to generate some dialogue, but the therapeutic suggestions are themselves constructed with symbolic methods \cite{kretzschmar2019can}. Serena stands out as one of the few platforms that relies primarily on the generative approach for both dialogue simulation and the direction of the counseling session. 

We can also differentiate mental health counseling systems according to the psychological methodologies they utilize. ELIZA was Joseph Weizenbaum's tongue-in-cheek replication of a stereo typical Rogerian therapist: \textit{And how did that make you feel?} Also called Person Centered Therapy (PCT), this methodology was chosen primarily because, with the technology of the time, it was relatively easy to mimic a Rogerian therapist by isolating keywords in the user prompt, and adding these keywords to a dictionary of predefined open-ended questions. Unclear or uncategorized keywords were handled with catch-all phrases. ELIZA is famous to this day because this crude approach had good results. Much to Weizenbaum's surprise, people actually enjoyed talking to his machine \cite{natale2019if}. 

Many contemporary digital mental health dialogue systems, including Woebot, Joy, and Wysa, primarily or substantially use Cognitive Behavioral Therapy (CBT). In addition to the method's proven efficacy in addressing a wide range of mental health challenges \cite{hoifodt2011effectiveness}, this therapy style lends itself well to being captured via symbolic methods, i.e. \textit{If user reports feeling X, recommend Y }. Serena stands out from its contemporaries by its choice to provide PCT to its users. PCT is an effective means of treating a plethora of mental health ailments \cite{murphy2016person}, and also provides the user with a totally open-ended and more life-like dialogue experience. We believe that the cutting edge of machine learning technology is especially well-suited to generating a person centered therapy session, that the user can direct in any way they choose. For example, it's possible to also interact with Serena outside of a therapeutic context simply by choosing another topic of conversation.  

\section{Methods and Models}

Serena consists of multiple natural language processing (NLP) models and heuristics that work together sequentially to obtain the most suitable responses to the input messages (\cref{overview}). The first stage consists of a large Seq2Seq Transformer \cite{vaswani2017attention} model which generates a beam of candidate responses. In the second stage a number of smaller, more specialized Transformer-based models and several heuristic rules are applied to select the final response from the candidate list.

\subsection*{Core Generative Model}

The generative model at the core of Serena is based on the pre-trained generative model described in \cite{roller2020recipes}. It is a 2.7 billion parameter Transformer architecture with 2 encoder layers, 24 decoder layers, 2560 dimensional embeddings, and 32 attention heads. The Transformer is a deep learning architecture that leverages the attention mechanism to provide contextual information for any position in the input sequence. This allows the Transformer to process the whole input sequence at once, in contrast to recurrent neural networks, which allows for a large increase in parallelization during training, increasingly making them the architecture of choice for NLP tasks. We are using the Transformer in the standard Seq2Seq setting, meaning that input sequences  are transformed into output sequences. In our case we transform a user prompt into the model's response.

The model has been pre-trained on the Pushshift Reddit Dataset which includes 651 million submissions and 5.6 billion comments on Reddit  between 2005 and 2019 \cite{baumgartner2020pushshift}. The objective of the pre-training was to generate a comment conditioned on the full thread leading up to a particular comment. 

We have fine-tuned this model on transcripts of counseling and psychotherapy sessions extracted from \cite{mcnally2014counseling}. Since during pre-training the context and response length was truncated at a length of 128 tokens we only use such samples for fine-tuning with the same maximum length. Our resulting counseling and psychotherapy dataset consisted of 14,300 patient's prompt and counselor's answer pairs. To perform the fine-tuning, we make use of the ParlAI \cite{miller2017parlai} platform using the default training parameters.

\subsection*{Beam Search and Post-Processing}

During decoding, ten candidate responses from the core generative model are processed and analyzed in a  so-called beam search. Since we have observed that the quality of candidate responses can vary wildly, despite being similarly ranked by the generative model, we employ additional processing to choose the most suitable response. In particular, we are using three pre-trained Transformer-based models for the following tasks: (1) detecting contradictions, (2) recognizing toxic language and (3) obtaining semantic sentence embeddings to detect repetitive answers. 

For the detection of contradictions we use a RoBERTa model \cite{liu2019roberta} pre-trained \cite{nie-etal-2020-adversarial} on several natural language inference datasets such as SNLI \cite{snli} and MNLI \cite{mnli}. Given two input sentences the model predicts three categories: contradiction, neutral and entailment. We use this model to detect whether a) sentences in a single candidate response and b) the user prompt and a candidate response are contradictory.

Since our generative model has been pre-trained on a large collection of Reddit comments, it may have been exposed to toxic language. To recognize toxic speech and exclude it from the model's responses we use the pre-trained ``unbiased'' model made available in the Detoxify repository \cite{Detoxify}. It is a RoBERTa model that has been trained on the Civil Comments \cite{civilcomments} dataset, a large collection of annotated comments with classes such as threat, insult or obscene. 

To obtain semantically meaningful sentence embeddings we are using a pre-trained SBERT model \cite{reimers-2019-sentence-bert} and calculate the cosine-similarity to estimate how similar a candidate response is to previous responses of the model. If the cosine-similarity is large, the response is considered to be repetitive and is rejected.

Additionally, we have curated a list of phrases that are undesirable and candidate responses containing them are avoided. Examples include phrases that are generally not helpful such as ``I don't know what to say'' or that are not considered to have any therapeutic value such as ``you just have to get over it''.

Finally, among the candidate responses that have not been rejected by any of the above post-processing steps we choose as the final response  the one that has been assigned the highest probability by the generative model.

\section{Results}
\subsection*{Deployment}
We have deployed the model using Google Kubernetes Engine (GKE) and it can be interacted with on our website\footnote{\url{https://serena.chat}}. Our implementation makes use of the ParlAI platform\footnote{\url{https://parl.ai}}, leveraging the abstractions it offers for setting up an interactive dialogue model. Our wbsite fetches responses from the model via REST API using FastAPI\footnote{\url{https://fastapi.tiangolo.com}}. For deployment with GKE the model had to be containerized and is running on a single Nvidia T4 GPU. 

Our deployment contains a survey that users can fill in after they have interacted with the model for some time. Users are queried to rate the degree to which the model understands their messages and whether they find the generated responses engaging and helpful.

\subsection*{Behaviors}
\begin{figure}[ht!]
\centering 
\includegraphics[width=0.4\textwidth]{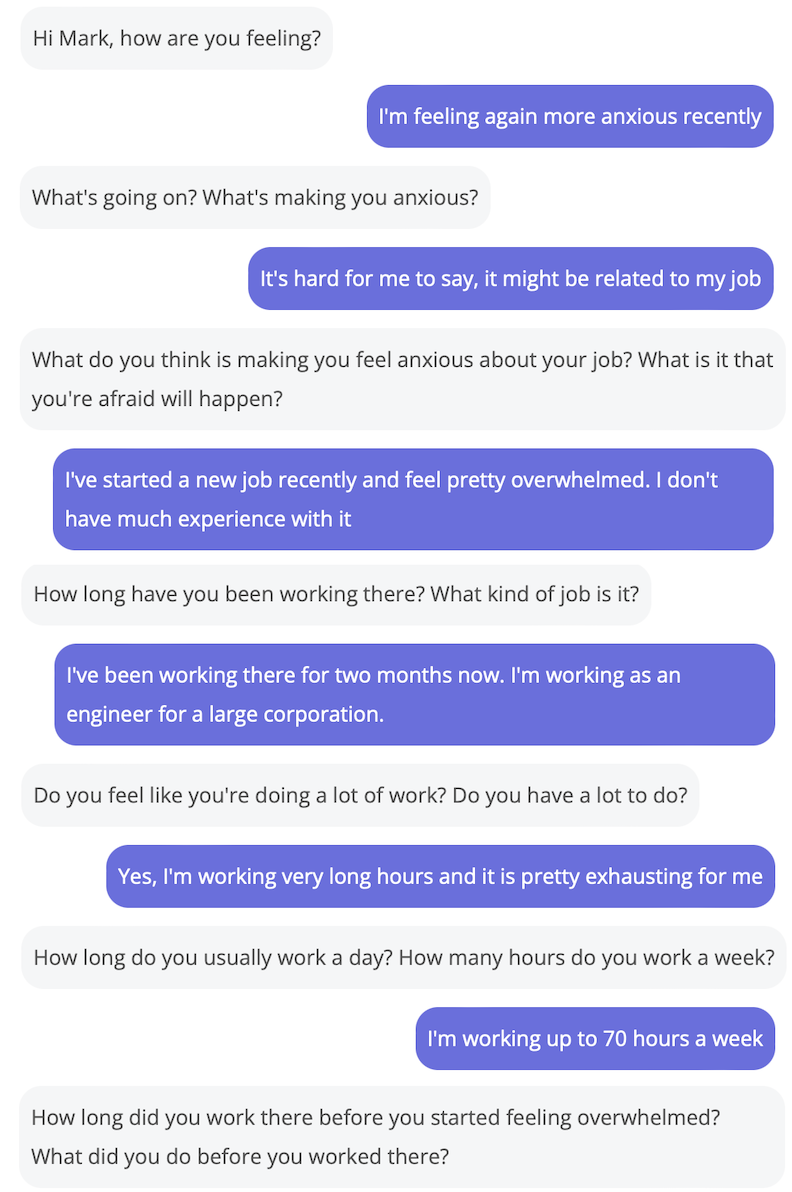}
\caption{Example dialogue from Serena.} 
\label{example2}
\end{figure}

Our dialogue model displays coherent understanding of the user's prompts and is capable of responding in a seemingly empathetic way (an example in \cref{example2}). The model is interacting with the user in an engaging way by asking relevant questions which encourages further introspection. 

One of the main issues with Serena is that it often hallucinates knowledge about the user, which is a well-known problem with transformer-based generative models \cite{hallucinate}. She will, for instance, claim that she has seen the user before or pretends to have detailed knowledge about their personal background. At the moment, we are trying to mitigate this issue by adding  phrases indicative of such hallucinations to the aforementioned exclude list. It has been hypothesized that hallucinations arise due to noise in the data, such as information in the output which can not be found in the input, and we plan to explore a possible solution proposed in \cite{filippova2020controlled}.

Another issue is that Serena tends to respond to the user's prompts only using questions. While this is desirable in terms of engaging the user in the conversation, early feedback from test users indicates that this behavior is perceived as annoying and even rude. Our current approach to this problem is to hard-code rules for choosing from the candidate responses such that the amount of generated questions is limited. This approach easily fails, however, since responses that are not questions are often missing in the candidate list. We plan to tackle this issue by carefully balancing the amount of questions and statements in the data used for fine-tuning the generative model.


\section{Discussion}

Recognizing that mental health care is often too expensive, too inconvenient, or too stigmatized, our team created a generative deep learning dialogue model that acts as a real-time companion and mental health counselor called \emph{Serena}. As described above, the combination of a core generative model with targeted post-processing leverages the Transformer architecture's potential \cite{vaswani2017attention} to exhibit natural language understanding and processing. In contrast to most of the available therapy chatbots focusing on CBT, Serena is designed to provide PCT focusing on a user's self-reflection and self-actualization \cite{rogers2012client}. In addition, as a generative deep learning model, Serena may contain potential bias, incoherence, and distaste, despite heuristics and post-processing attempts to mitigate such responses. Nonetheless, our generative model has the potential to address the accessibility problem of mental health counseling.

Serena addresses the issue of cost because it is inherently less expensive to use than a human counselor \cite{pricehumand}. 

As far as time is concerned, Serena also presents advantages compared to traditional counseling. Serena can counsel its users' wherever they choose, so it negates the need to commute or to make changes to the users daily routine and responsibilities. 

Finally, Serena removes most of the fear and stigma associated with traditional mental health therapy methods. While human users are apt to anthropomorphize virtual dialogue interfaces \cite{weizenbaum1976computer}, there is scant evidence that it brings along with it any of the fear of shyness that can arise from interaction with a human interlocutor.

In future works, we further hope to improve the model and the UX/UI through focus groups of psychologists and clients. With our built-in survey, we plan to substantiate our internal testing of the model's behavior with large-scale results from our survey. Our internal testing is based on our growing database of prompt and responses, which allows us to not only fine-tune our model, but to understand the needs and behaviors of our users. We also recognize the upmost important of respecting our users' privacy, and to this end we have encrypted all the user data generated on our platform, and we empower users to select exactly which data they chose to share with us. By protecting the privacy of our users, we give them the peace of mind to use Serena just as we intended: a trusted confidante always at their fingertips in a time of need. 

\section*{Acknowledgments}
This research was carried out with the support of the Interdisciplinary Centre for Mathematical and Computational Modelling University of Warsaw (ICM UW) under computational allocation no GDM-3540; the NVIDIA Corporation’s GPU grant; and the Google Cloud Research Innovators program. LB and NCC are partially supported by by National Science Centre (NCN) of Poland [2020/02/Y/ST6/00071], the CHIST-ERA grant [CHIST-ERA-19-XAI-007].

\bibliographystyle{ieee_fullname}
\bibliography{refs}

\end{document}